\def\year{2018}\relax
\documentclass[letterpaper]{article} 
\usepackage{aaai18}  
\usepackage{amsmath}
\usepackage{times}  
\usepackage{helvet}  
\usepackage{courier}  
\usepackage{url}  
\usepackage{graphicx}  
\usepackage[position=b]{subfig}
\usepackage{caption}
\usepackage{algorithm}
\usepackage{algpseudocode}
\usepackage{multirow}
\usepackage{booktabs}

\begin{document}
%
\title{AdaComp : Adaptive Residual Gradient Compression for \\Data-Parallel Distributed Training {}}
\author{Chia-Yu Chen, Jungwook Choi, Daniel Brand, Ankur Agrawal, Wei Zhang, Kailash Gopalakrishnan\\
IBM Research AI\\
1101 Kitchawan Rd. Yorktown Heights, New York 10598\\
\{cchen, choij, danbrand, ankuragr, weiz, kailash\}@us.ibm.com
}

\maketitle
\begin{abstract}
Highly distributed training of Deep Neural Networks (DNNs) on future compute platforms (offering 100 of TeraOps/s of computational capacity) is expected to be severely communication constrained. To overcome this limitation, new gradient compression techniques are needed that are computationally friendly, applicable to a wide variety of layers seen in Deep Neural Networks and adaptable to variations in network architectures as well as their hyper-parameters. In this paper we introduce a novel technique - the Adaptive Residual Gradient Compression (\textbf{AdaComp}) scheme. AdaComp is based on localized selection of gradient residues and automatically tunes the compression rate depending on local activity. We show excellent results on a wide spectrum of state of the art Deep Learning models in multiple domains (vision, speech,  language), datasets (MNIST, CIFAR10, ImageNet, BN50, Shakespeare), optimizers (SGD with momentum, Adam) and network parameters (number of learners, minibatch-size etc.). Exploiting both sparsity and quantization, we demonstrate end-to-end compression rates of ${\sim}200\times$  for  fully-connected  and  recurrent  layers, and ${\sim}40\times$ for convolutional layers, without any noticeable degradation in model accuracies.
\end{abstract}

\section{Introduction}
Over the past decade, Deep Learning (DL) has emerged as the dominant Machine Learning algorithm showing remarkable success in a wide spectrum of application domains ranging from image processing \cite{he2016deep}, machine translation \cite{wu2016google}, speech recognition \cite{xiong2017microsoft} and many others. In each of these domains, DNN’s achieve superior accuracy through the use of very large and deep models - necessitating up to 100’s of ExaOps of computation during training (on GPUs) and GB’s of model and data storage.

In order to improve the training time over single-node systems, distributed algorithms \cite{chilimbi2014project}, \cite{ho2013more}, \cite{lian2015asynchronous} are frequently employed to distribute the training data over multiple CPUs or GPUs - using data parallelism \cite{gupta2016model}, model parallelism \cite{dean2012large} and pipeline parallelism \cite{wu2016google} techniques. Data parallelism techniques are widely applied to distribute convolutional layers while model and pipeline parallelism approaches are more effective for fully-connected and recurrent layers of a neural net. All 3 techniques necessitate very high interconnect bandwidth between the GPUs (in order to communicate the necessary parameters) and impose limits on peak system utilization and model training time. A variety of system topologies have also been explored to efficiently implement distributed training using data parallelism techniques ranging from simple parameter-server based approaches to the recently proposed Wild-Fire technique \cite{raviWildfire} that allows direct exchange of weights during reduction. Recently, ring-based system topologies \cite{fastmultiGPU} have been proposed to maximally utilize inter-accelerator bandwidths by connecting all accelerators in the system in a ring network. The accelerator then transports it’s computed weight gradients (from its local mini-batch) directly to the adjacent accelerator (without the use of centralized parameter servers). However, as the number of learners increases, distribution of the minibatch data under strong scaling conditions has the adverse effect of significantly increasing the demand for communication bandwidth between the learners while proportionally decreasing the FLOPs needed in each learner - creating a severe computation to communication imbalance.

Simultaneously, there has been a renaissance in the computational throughput (TeraOps per second) of DL training accelerators - with accelerator throughputs exceeding 100’s of TeraOps/s expected in the next few years \cite{voltaGPU}, \cite{googleTPU}. Exploiting hardware architectures based on reduced precision \cite{gupta2015deep}, \cite{courbariaux2014training}, these accelerators promise dramatic reduction in training times in comparison to commercially available GPUs today. For a given DL network and distribution approach, the bandwidth needed for inter-accelerator communication (in GB/s) scales up directly with raw hardware performance as well as the number of learners. In order to guarantee high system performance, radically new compression techniques are therefore needed to minimize the amount of data exchanged between accelerators. Furthermore, the time required for compression needs to be significantly smaller than the computational time required for back-propagation. Section 2 discusses various prior compression techniques - most of which were applied to Fully Connected (FC) layers of DL networks. But high computational capacity and wide distribution necessitates compression of convolutional layers in addition to the fully-connected ones. Extending this need to networks that have a mix of fully-connected, convolutional and recurrent layers, it is desirable to have a universally applicable and computationally-friendly compression scheme that does not impact model convergence and has minimal new hyper-parameters. In this paper, we propose a new gradient-weight compression scheme for distributed deep-learning training platforms that fully satisfies these difficult constraints. Our primary contributions in this work include:
\begin{itemize}
\item We explore the limitations of current compression schemes and conclude that they are not robust enough to handle the diversity seen in typical neural networks.
\item We propose a novel computationally-friendly gradient compression scheme, based on simple local sampling, called AdaComp. We show that this new technique, remarkably, self-adapts its compression rate across mini-batches and layers.
\item We also demonstrate that the new technique results in a very high net compression rate ( ${\sim}200\times$ in FC and LSTM layers and ${\sim}40\times$ in convolution layers), with negligible accuracy and convergence rate loss across several network architectures (CNNs, DNNs, LSTMs), data sets (MNIST, CIFAR10, ImageNet, BN50, Shakespeare), optimizers (SGD with momentum, ADAM) and network-parameters (mini-batch size and number of learners).
\end{itemize}

\section{Residual gradient compression}
\subsection{Background}
Given the popularity of distributed training of deep networks, a number of interesting techniques for compressing FC weight gradients have been proposed \cite{seide20141}, \cite{strom2015scalable}, \cite{dryden2016communication}. Siede \cite{seide20141} proposed a one-bit quantization scheme for gradients, which locally stores quantization errors, and reconstruction values are computed using the mean of the gradient values. This scheme achieves a fixed compression rate of 32x applicable only to FC layers. Strom \cite{strom2015scalable} proposed a thresholding technique for FC layers that is somewhat similar in approach but can provide much higher compression rates. Only gradient values that exceed a given threshold are quantized to one bit and subsequently propagated to the parameter server along with their indices. These schemes preserve quantization error information in order to reduce the impact of thresholding on the accuracy of the trained model. Furthermore, these papers do not discuss techniques for determining an optimal threshold value. Dryden \cite{dryden2016communication} proposed a ``best of both worlds'' approach, by combining the one-bit quantization and thresholding ideas. Instead of using a fixed threshold, they propagate a fixed percentage of the gradients, and use the mean of the propagated values for reconstruction. This technique requires sorting of the entire gradient vector which is a computationally expensive task, particularly on a special-purpose accelerator. All the three techniques above are aimed at reducing weight update traffic in deep multi-layer perceptrons composed of fully-connected layers. One common principle that allows these compression techniques to work without much loss of accuracy is that each learner maintains an accumulated gradient (that we refer to as \textbf{residual gradients}) comprising of the gradients that have not yet been updated centrally. We exploit a similar principle in our work.

\subsection{Related work}
For convolutional neural networks (CNNs), the primary motivation for reducing the size of the network has stemmed from the desire to have small efficient models for inference and not from training efficiency point of view. Han et al. \cite{han2015deep} combine network pruning along with quantization and Huffman encoding to decrease the total size of the weights in the model. This is useful as a fine-tuning technique after a trained network is available. Similarly, Molchanov \cite{molchanov2016pruning} presented a technique for pruning feature maps while adapting a trained CNN model for distinct tasks using transfer learning.

Recently, a ternary scheme was proposed \cite{wen2017terngrad} to compress communicated gradients while training CNNs for image processing. Without the use of sparsity, the compression rate in their approach is limited to $16\times$ for both fully connected and convolutional layers. Much higher compression rates (${\sim}100\times$) will be needed in the future to balance computation and communication for emerging deep learning accelerators \cite{voltaGPU}. Furthermore, their technique showed significant (${\sim}1.5\%$) degradation in large networks including GoogLeNet. In our work, we apply a novel scheme that exploits both sparsity and quantization to achieve much higher end-to-end compression rates and demonstrate that \textless 1\% degradation is achievable in many state of art networks (e.g. ResNet50 for ImageNet) as well as other application domains (including speech and language). In addition, our approach focuses on compression, which is complimentary to the techniques used by Goyal \cite{goyal2017accurate} and Minsik \cite{cho2017powerai} on their use of large minibatches for highly distributed training of deep networks.

\subsection{Observations}
We first note that most of the residual gradient compression techniques discussed in Background section do not work well for convolutional layers during training - especially when we also compress the fully connected layers of the same model. As Fig. 1 shows, even for a simple dataset like CIFAR10 \cite{krizhevsky2009learning} \cite{jia2014caffe}, when the fully connected layer is compressed, a simple 1-bit quantization based compression scheme \cite{seide20141} for the convolutional layer significantly worsens model accuracy. What is desired is a universal technique that can compress all kinds of layers (including convolutional layers) without degrading model convergence.

\begin{figure}[h]
  \centering
  \includegraphics[height=43.5mm]{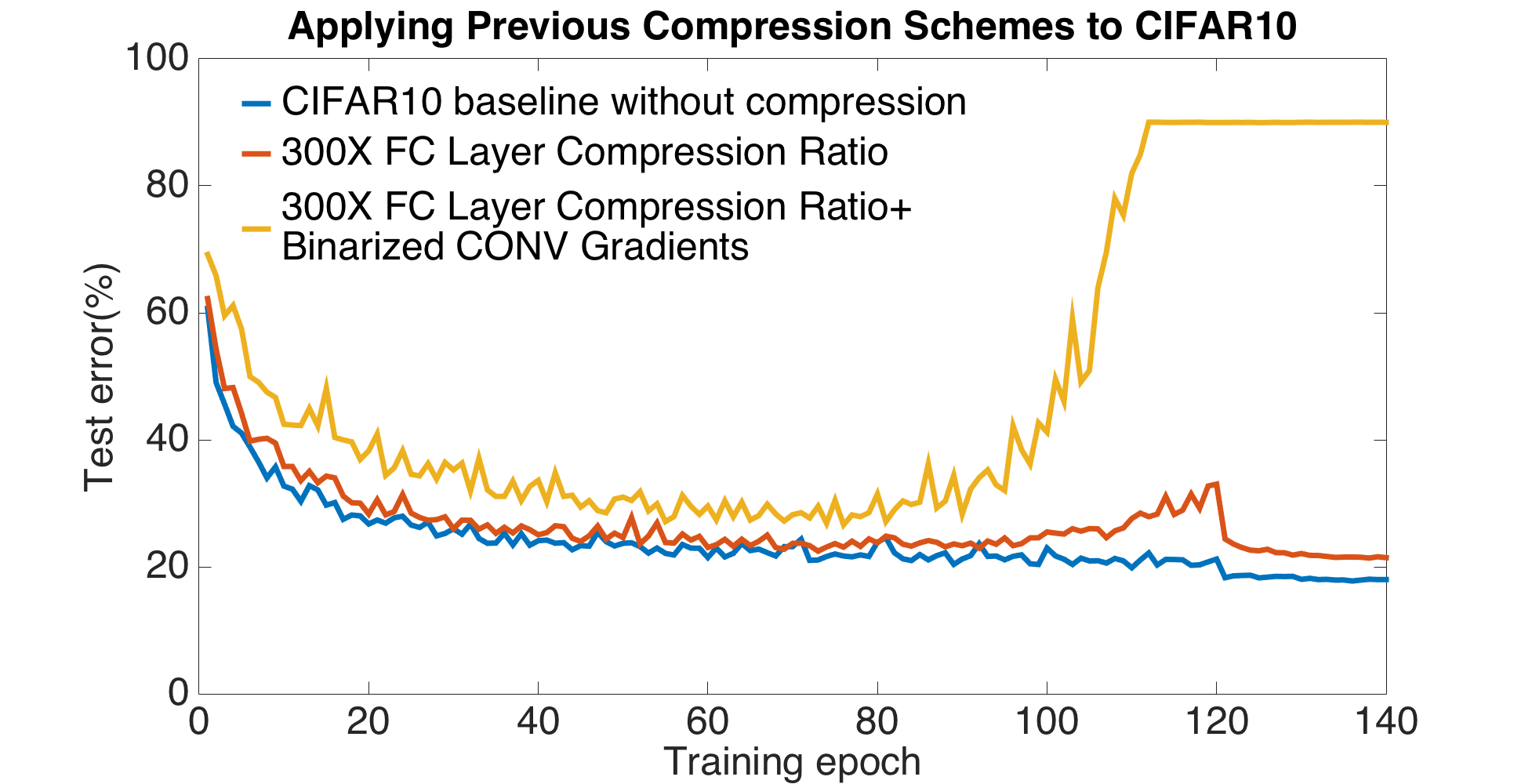}
  \caption{For the CIFAR10 dataset and a network similar to the one in Caffe \cite{jia2014caffe}, compressing the FC layer only by sending the top 0.3\% of the gradients \cite{dryden2016communication} results in a modest degradation in test accuracy (20\% vs. 18\%). Furthermore, additionally compressing the convolutional layer using a 1-bit quantization scheme \cite{seide20141} results in complete model divergence.}
\end{figure}

Our first observation, inspired by \cite{molchanov2016pruning}, is that the best pruning techniques consider not only the magnitude of the weights, but also the mini-batch data (and their impact on input feature activities) to maximize pruning efficiency. However, prior work in the compression space assumes that the magnitude of the residue is the only metric indicative of its overall importance. For example, in \cite{strom2015scalable} a fixed gradient value is used as the threshold, and in \cite{dryden2016communication} a fixed portion of the top gradients are communicated. Intuitively, picking residues with maximum value for any layer may miss lower-magnitude (but critically important) residual gradients that are connected to high activity input features. We also note (empirically) that accumulated residual gradients show little to no correlation with feature activations - which raises further questions about the efficacy of simply picking high valued gradients in any layer.

The second key observation is that neural networks have different activities in different layers depending on overall network architecture, layer types, mini-batch sizes and other parameters, and all of these factors have a direct impact on compression rate. Since compression during training has the potential to dramatically affect convergence, we note that new techniques are desired that automatically and universally adapt to all of these variations without necessitating a whole new set of hyper-parameters.

Finally, it is critical to minimize the computational overhead for any new compression technique given the dramatic reduction in the training time per mini-batch with new high TeraOp/s accelerators. This constraint automatically precludes techniques that require globally sorting of the residue vectors, and instead drives us towards techniques that are accelerator friendly and localized (in terms of memory accesses).

\subsection{Adaptive Residual Gradient Compression (AdaComp) Technique}
Given the lack of correlation between the activity of the input features and the residual gradients in any layer, we conjecture that it is important to have a small enough sampling window that can effectively capture the right residues across the entire layer. To facilitate this, we divide the entire residue vector for each layer (laid out as $NumOutMaps\times NumInpMaps\times KernelRows\times KernelCols$) uniformly into several bins - where the fixed length bin size, {${\rm \textbf L}_{\rm \small{\textbf T}}$} is a new hyper-parameter. In each bin, we first find the maximum of the absolute value of the residues. In addition to this value, we found that it was also important to send several other residues that are relatively similar in magnitude to this maximum. There are several ways to find such important gradients inside each bin. In this paper, we propose a relatively simple self-adjusting scheme. Recall that in each mini-batch, the residue is computed as the sum of the previous residue and the latest gradient value obtained from back-propagation. If the sum of its previous residue plus the latest gradient multiplied by a scale-factor exceeds the maximum of the bin, we include these additional residues in the set of values to be sent (and centrally updated). Empirically, we studied a range of choices for the scale factor (from 1.5 - 3.0$\times$) and chose 2$\times$ primarily for computational ease (simple additions vs. multiplications). The primary intuition here is that since the residues are empirically much larger than the gradients, this scheme allows us to send a whole list of important residues close to the {local maximum}. Furthermore, we quantize the compressed residue vector in order to increase the overall compression rate. AdaComp is applied to every layer separately - and each learner sends a scale-factor in addition to the compressed sparse vector.

This approach to ``threshold'' the selection is self-adjusting in 3 ways. First, it allows some bins to send more gradients than others - as many as are needed to accurately represent each bin. Secondly, since the residues are small in the early epochs, more gradients are automatically transmitted in comparison with later epochs. Third, as will be shown in later sections, in comparison to other schemes, this technique minimizes the chances of model divergence that result from an explosion in the residual gradient values and gradient staleness. Thus, AdaComp adaptively adjusts compression ratios in different mini-batches, epochs, network layers and bins. These characteristics provide automatic tuning of the compression ratio, resulting in robust model convergence. We observe that just one hyper-parameter (\textbf{${\rm L}_{\rm T}$}) is sufficient to achieve high compression rates without loss of accuracy.
Finally, it should also be noted that AdaComp, unlike \cite{dryden2016communication}, does not require any sorting (or approximations to sorting) and is therefore computationally very efficient (O(N)) for high-performance systems.

\subsection{Pseudo code}
\label{subsec:Pseudo-code}
The following pseudo code describes two algorithms. Algorithm 1 shows the gradient weight communication scheme we used to test AdaComp, and algorithm 2 is the AdaComp algorithm we propose. Note that our algorithm is not limited to a particular quantization function. In this work, the quantization function uses a sign bit and a scale value to represent the original number. In addition, we use a single scale value for the entire layer - calculated as the average of the absolute values of all the elements in the ${\rm \text g}_{\rm \small{\text max}}$ vector. For simplicity of exposition Algorithm 2 assumes that the threshold T evenly divides the $length(G)$.

\begin{algorithm}
\caption{Computation Steps}
    \begin{algorithmic}
        \State $\texttt{\small{learningNoUpdate()}}$     \Comment{\small{Forward/Backward only} }
        \State $\texttt{serializeGrad()}$  \Comment{Collect grad of each layer as a vector}
        \State $\texttt{pack()}$                 \Comment{AdaComp Compression for each layer}
        \State $\texttt{exchange()}$\Comment{Learner receives packed grads from others}
        \State $\texttt{unpack()}$               \Comment{AdaComp Decompression for each layer}
        \State $\texttt{averageGradients()}$            \Comment{Average among all learners}         
        \State $\texttt{updateWeights()}$                 \Comment{Performed locally at each learner}
    \end{algorithmic}
\end{algorithm}

\begin{algorithm}
\caption{Details of pack()}
\label{alg:Details-of-pack}
    \begin{algorithmic}
        \State\small{$G\gets  residue + dW$} \Comment{\small{dW is from \texttt{serializeGrad()}}}
        \State$H\gets  G + dW$ \Comment{\small{H = Residue + 2*dW}}
        \State$\textit{Divide G into bins of size T}$
        \For{$i \gets 1, length(G)/T$} \Comment Over all bins 
            \State $\textit{Calculate }g_{max}(i);$ \Comment Get largest absolute value in each bin
        \EndFor 
        \For{$i \gets 1, length(G)/T$} \Comment Over all bins
            \For{$j \gets 1,T$} \Comment Over all indices within each bin
                \State$index\gets (i-1)*T+j$ 
                \If{$\mid H(index)\mid  \geq g_{max}(i)$} \Comment Local max compare
                    \State$Gq(index)\gets \textit{Quantize(G(index))}$
                    \State$\textit{add Gq(index) to a pack vector (sent in exchange())}$
                    \State$residue(index)\gets G(index)-Gq(index)$ 
                \Else
                    \State$residue(index)\gets G(index)$  \Comment No transmission
                \EndIf
            \EndFor    
        \EndFor    
    \end{algorithmic}
\end{algorithm}

\section{Experiments}
We performed a suite of experiments using the AdaComp algorithm. That algorithm is encapsulated within two functions, pack() and unpack(), inserted into the standard DL flow between the backward pass and the weight-update step (Algorithm 1). The pack/unpack algorithms are independent of the exchange() function which depends on the topology (ring-based vs. parameter-server based), and therefore the exchange() function is not a subject of this paper. 
AdaComp impacts 4 critical parameters during DL training: 1) communication overheads, 2) extra time spent executing the pack() and unpack() functions, 3) convergence, and 4) compression ratios. In this paper, we evaluate convergence and compression (items 3 and 4) - but do not report the impact on run-time (items 1 and 2).

\subsection{Methodology}
Experiments were done using IBM SoftLayer cloud servers where each server node is equipped with two Intel Xeon E5-2690-V3 processors and two NVIDIA Tesla K80 cards. Each Xeon processor has 12 cores running at 2.66GHz and each Tesla K80 card contains two K40 GPUs each with 12GB of GDDR5 memory.
The software platform is an in-house distributed deep learning framework (\cite{gupta2016model}, \cite{raviWildfire}). The exchange of gradients is done in a peer-to-peer fashion using MPI. In addition, we use synchronous SGD - where all the learners always have identical weights at each step.

Table~\ref{tab:data_model} records the details of the datasets and neural network models we use in this paper. Here we briefly describe the network architectures used in our experiments.

\begin{itemize}
\item MNIST-CNN \cite{lecun1998gradient}: 2 convolutional layers (with 5x5 filters and Relu activation functions), 2 FC layers, and a 10-way softmax.
\item CIFAR-CNN \cite{krizhevsky2009learning}: 3 convolutional layers (with 5x5 filters and Relu activation function), 1 FC layer, and a 10-way softmax. 
\item AlexNet \cite{krizhevsky2012imagenet}: 5 convolutional layers and 3 FC layers. The output layer is a 1K softmax layer.
\item ResNet18 \cite{he2016deep}: 8 ResNet blocks totaling 16 convolutional layers with 3x3 filters, batch normalization, Relu activation and a final FC layer with a 1K softmax. 
\item ResNet50 \cite{he2016deep}: 16 bottleneck ResNet blocks totaling 48 convolutional layers with 3x3 or 1x1 filters, batch normalization, Relu activation and a final FC layer with a 1K softmax. 
\item BN50-DNN \cite{van2017training}: 6 FC layers (440x1024, 1024x1024, 1024x1024, 1024x1024, 1024x1024, 1024x5999) and a 5999-way softmax.
\item LSTM \cite{multiRNN}: 2 LSTM layers (67x512, 512x512), 1 FC layer (512x67) and a 67-way softmax.
\end{itemize}

\begin{table}[]
\small
\centering
\caption{Dataset and Model}
\label{tab:data_model}
\begin{tabular}{@{}lcclcl@{}}
\toprule
\multicolumn{3}{c}{Dataset}                                                                & \multicolumn{2}{c}{Model}       &  \\ 
\multicolumn{1}{c}{Name}  & \multicolumn{1}{l}{\#Sample} & Size                   & \multicolumn{1}{c}{Name} & Size &  \\ \midrule
MNIST                       & 60k                                     & 181MB
& CNN                      & 1.7MB    &  \\ \midrule
CIFAR10                       & 50k                                     & 1GB
& CNN                      & 0.3MB    &  \\ \midrule
\multirow{3}{*}{ImageNet } & \multirow{3}{*}{1.2M}              & \multirow{3}{*}{140GB} & AlexNet                 & 288MB    &  \\
                                                                                         &&& ResNet18                & 44.6MB      & \\
                                                                                         &&& ResNet50                & 98MB      & \\ \midrule
BN50                       & ~16M                                     & 28GB                      & DNN                      & 43MB    &  \\ \midrule
Shakespeare                 & 50k                                     & 4MB                      & LSTM                   & 13MB    &  \\ \bottomrule
\end{tabular}
\end{table}

\begin{table*}[ht!]
\small{
  \caption{CNN, MLP, and LSTM results }
  \label{sample-table}
  \centering
  \begin{tabular}{llllllll}
    \toprule
    \multicolumn{8}{c}{Compression hyper-parameters: convolution layer ${\rm L}_{\rm T}$: 50 and fully connected layer ${\rm L}_{\rm T}$: 500}                   \\
    \cmidrule{1-8}
    Model     & MNIST-CNN  & CIFAR10-CNN   & AlexNet     & ResNet18 & ResNet50  & BN50-DNN & LSTM \\
    \midrule
    Dataset   & MNIST & CIFAR10  & ImageNet  & ImageNet & ImageNet   & BN50 & Shakespeare    \\
    Mini-Batch size   & 100 & 128  & 256 & 256   & 256 & 256 & 10   \\
    Epochs    & 100 & 140 & 45 & 80 & 75   & 13 & 45     \\
    Baseline (top-1)    & 0.88\%   & 18\%  & 42.7\% & 32.41\% & 28.91\%  & 59.8\% & 1.73\%   \\
    Our method (top-1)  & 0.85\% (8L) & 18.4\%(128L) & 42.9\%(8L) & 32.87\%(4L) & 29.15\%(4L)  & 59.8\% (8L) & 1.75\% (8L)    \\
    Learner number     & 1,8  & 1,8,16,64,128  & 8 & 4 & 4  & 1,4,8 & 1,8    \\
    \bottomrule
  \end{tabular}
  }
\end{table*}

\begin{figure*}[htb!]
  \centering
        \subfloat[CIFAR10-CNN for CIFAR10 dataset: For Stress test under Extreme Compression, \textbf{${\rm L}_{\rm T}$} = 800 is used for CONV and 8000 is used for FC layers]{\includegraphics[height=43.5mm]{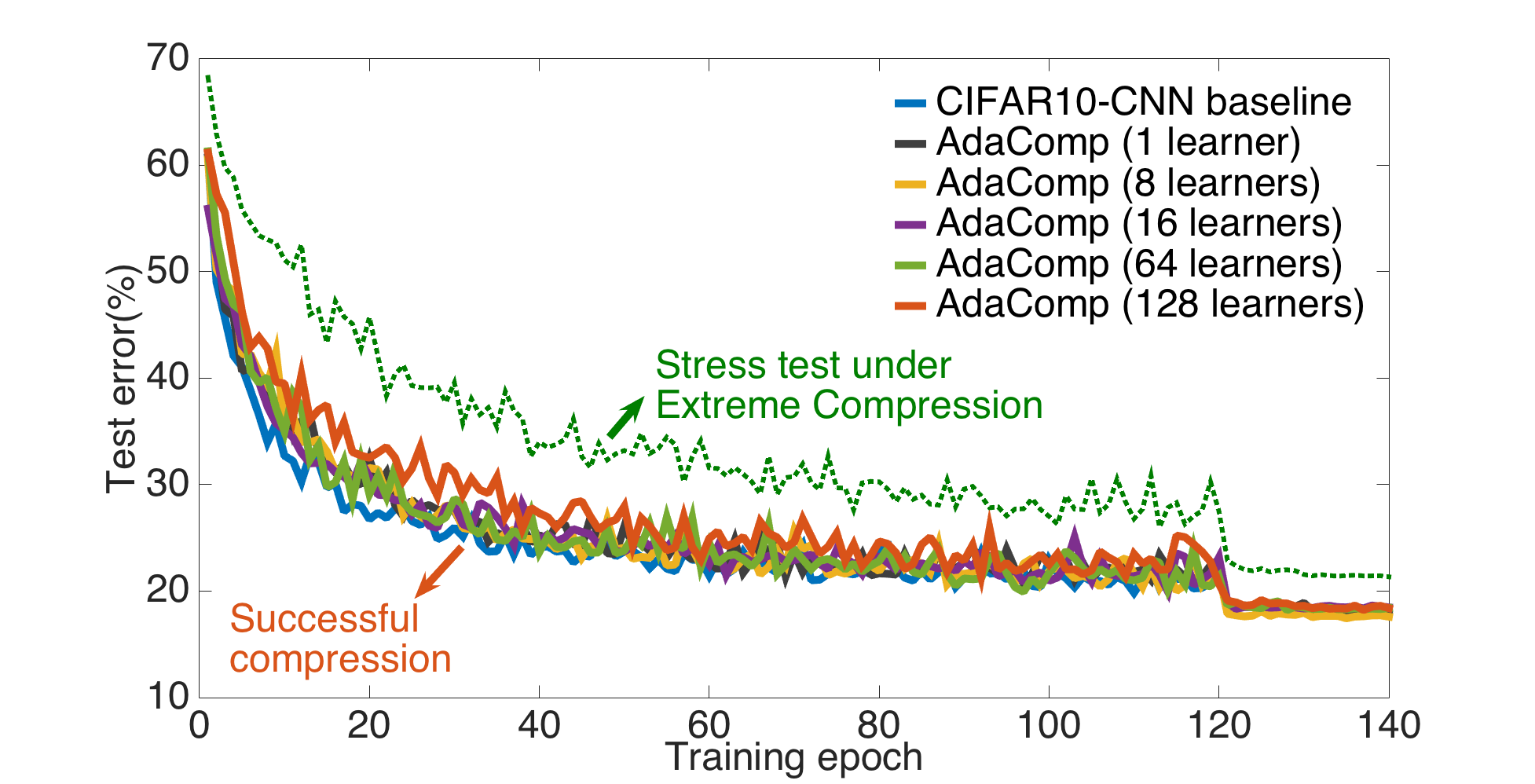}} \hspace{0.192cm}
        \subfloat[AlexNet for ImageNet dataset: For Stress test under Extreme Compression, \textbf{${\rm L}_{\rm T}$} = 500 was used for both CONV and FC layers.]{\includegraphics[height=43.5mm]{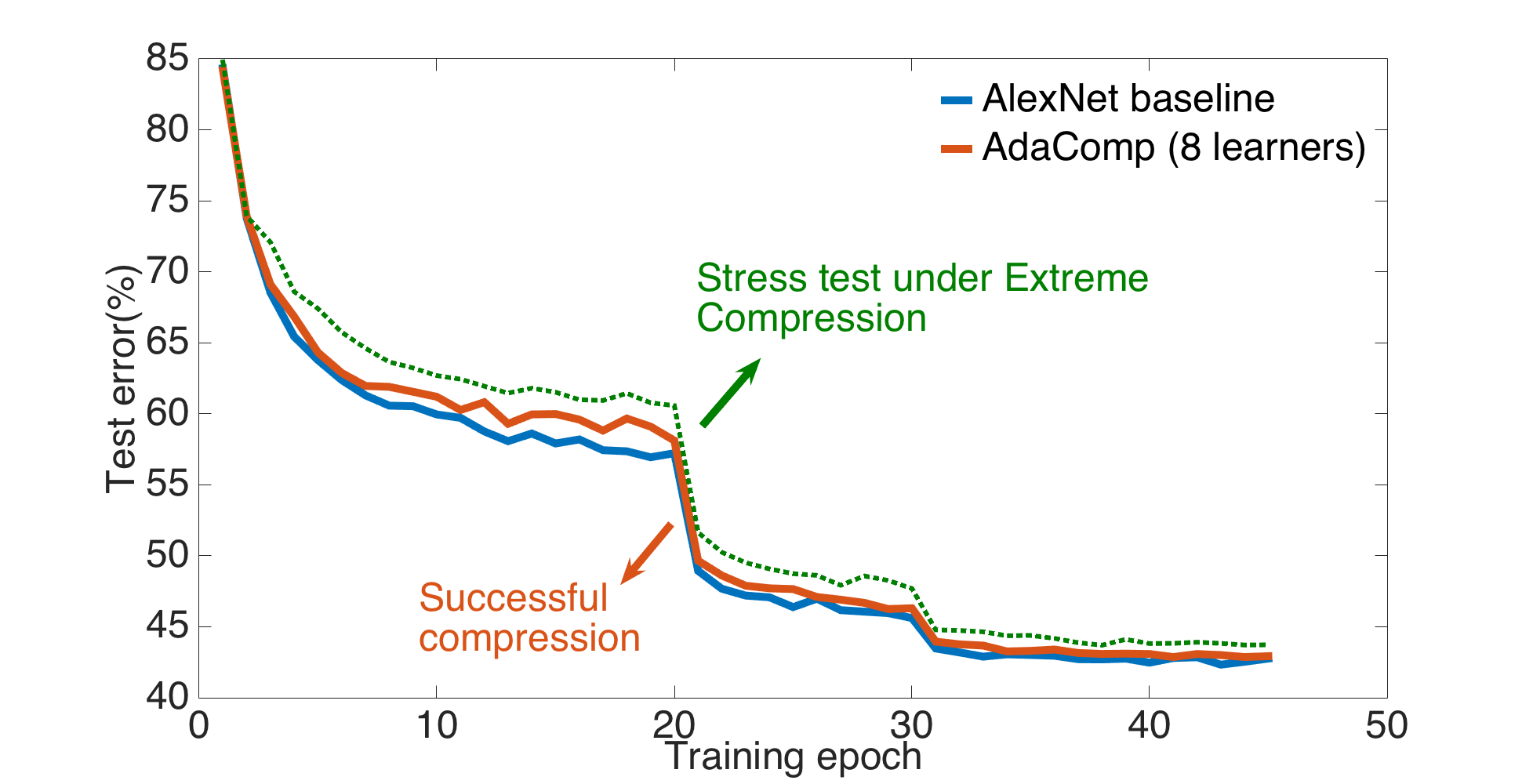}}

        \subfloat[ResNet18 for ImageNet dataset]{\includegraphics[height=43.5mm]{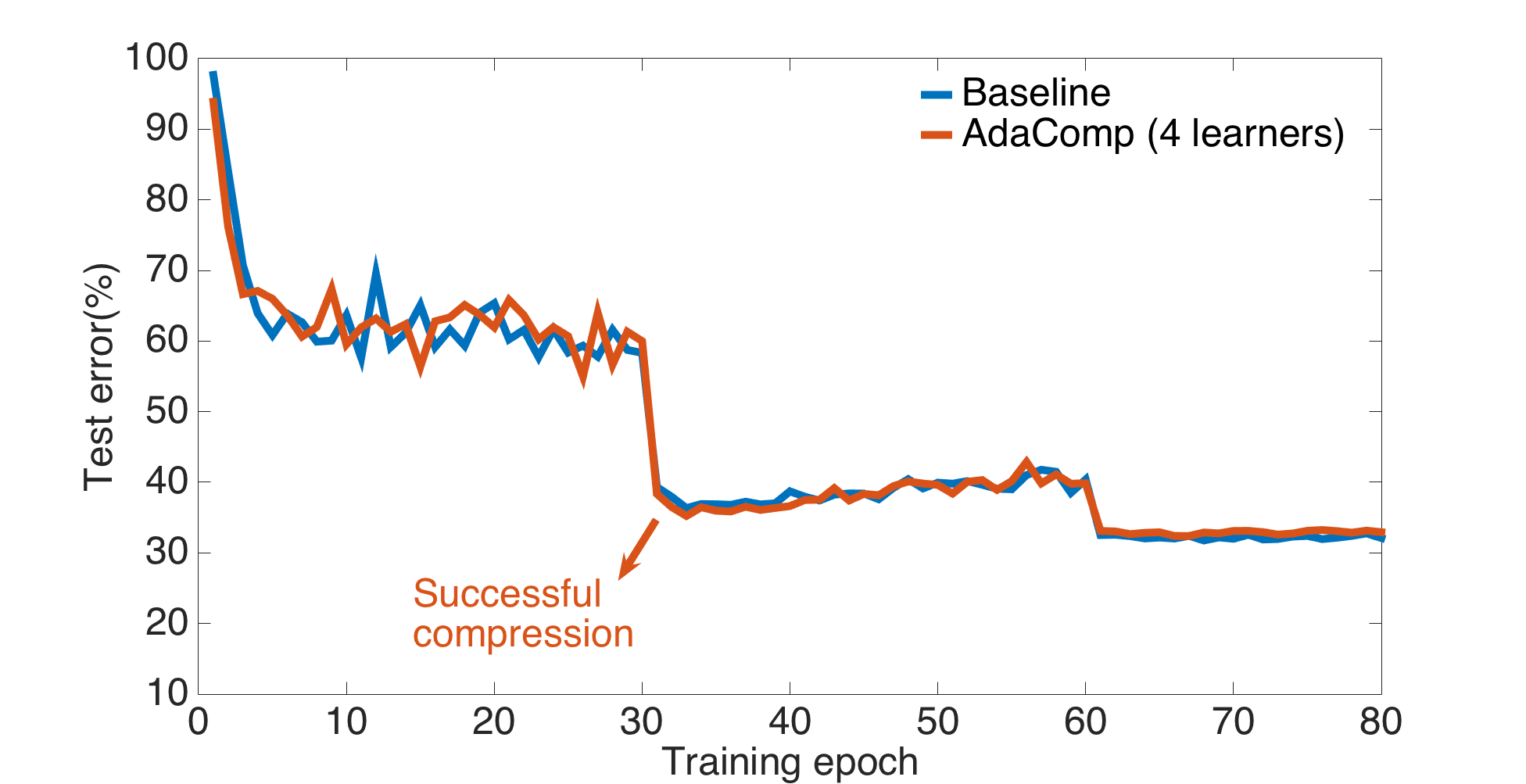}}
        \subfloat[ResNet50 for ImageNet dataset]{\includegraphics[height=43.5mm]{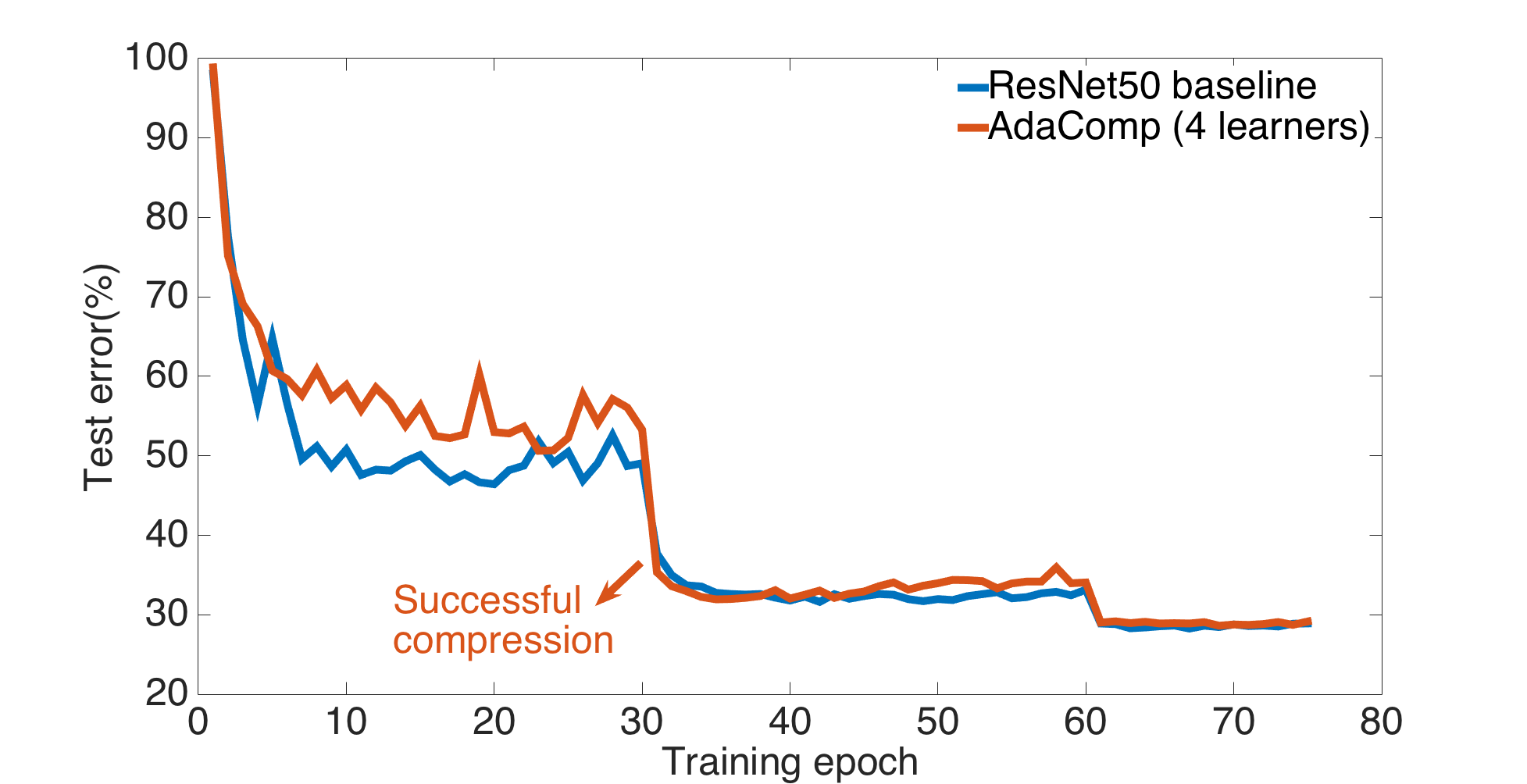}}

        \subfloat[LSTM for Shakespeare dataset]{\includegraphics[height=43.5mm]{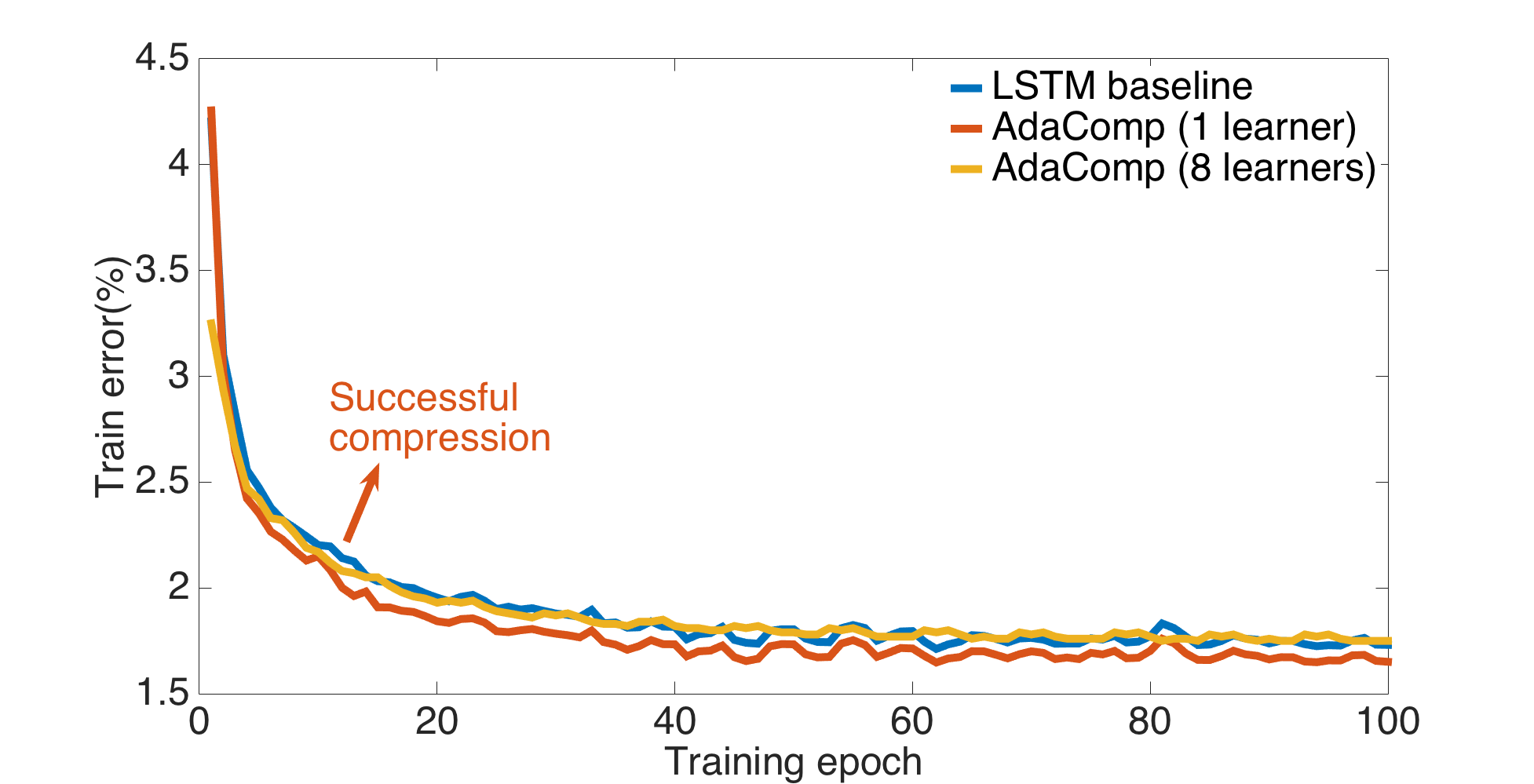}}
        \subfloat[BN50-DNN for BN50 speech dataset]{\includegraphics[height=43.5mm]{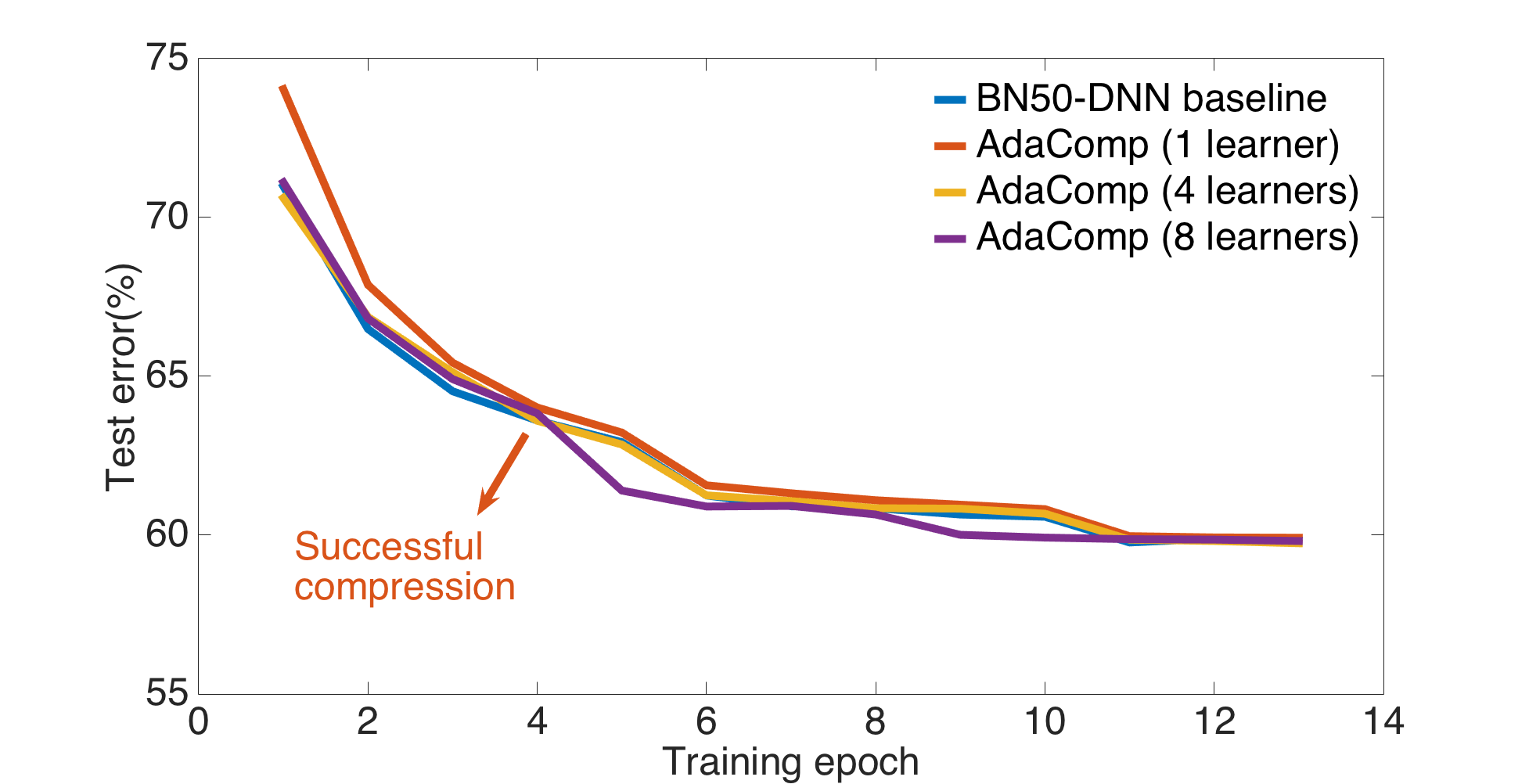}}

        \caption{{Model convergence results for different networks, datasets and learner numbers. In all tests (except for the Stress Tests) the same compression hyper-parameters are used: ${\rm L}_{\rm T}$ of 50 for CONV layers and 500 for FC/LSTM layers. Excellent compression ratios of $40\times$ for CONV layers and $200\times$ for FC/LSTM layers are obtained with no degradation in model convergence or accuracy. Stress tests are also shown for CIFAR10-CNN and AlexNet to demonstrate the limits of compression.
        }}
\end{figure*}

\begin{figure}[h]
  \centering
  \includegraphics[height=4.4cm]{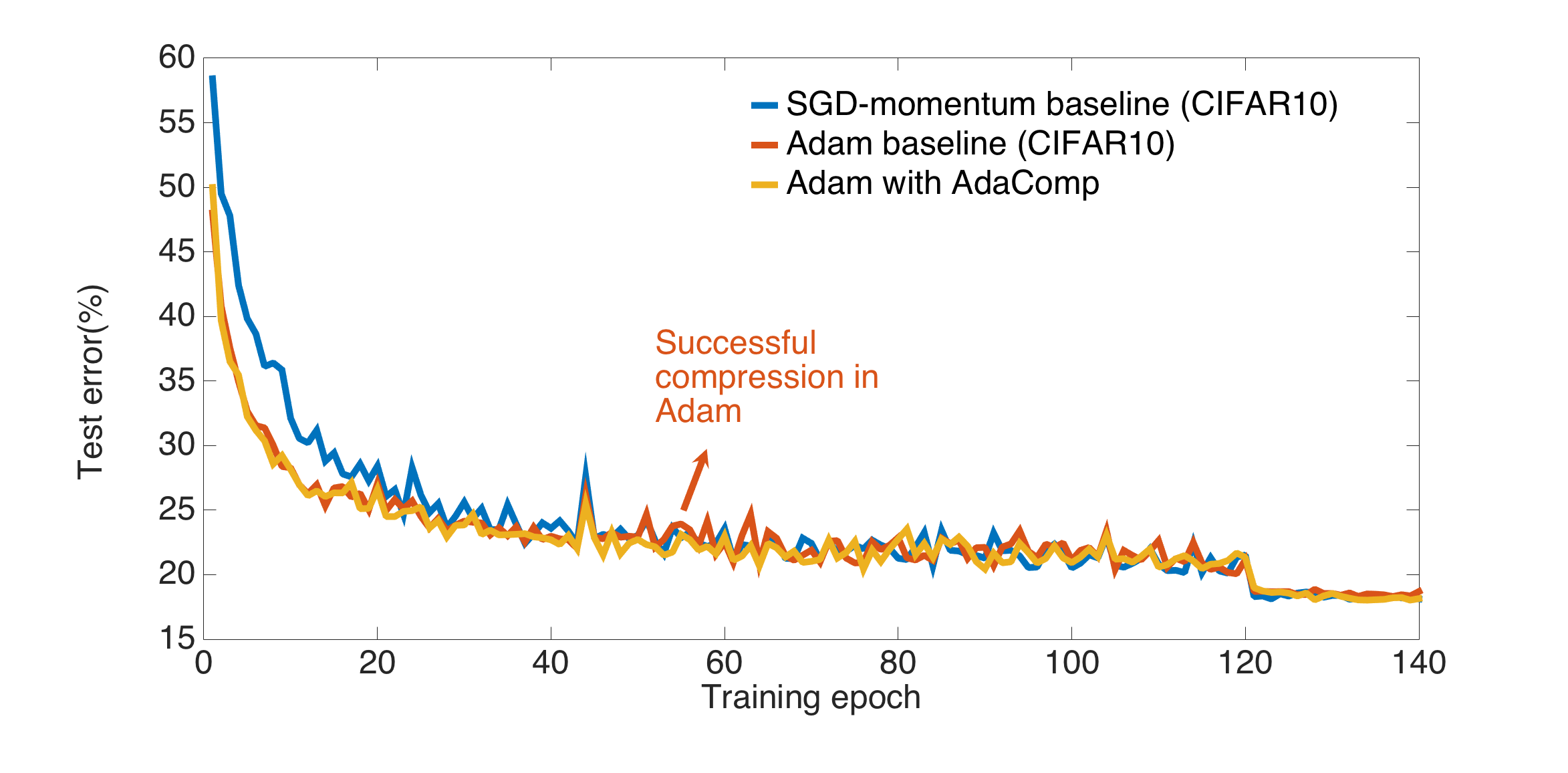}
  \caption{This work (AdaComp) achieves similar effective compression rates ($40\times$ for CONV layers and $200\times$ for FC/LSTM layers) with Adam, and had no impact on convergence or test error (Adam: baseline:18.1\% vs compression:18.3\%). In comparison to SGD, Adam exhibited faster initial convergence but similar final accuracy.}
\end{figure}

\subsection{Experiment Results}
To demonstrate the robustness as well as the wide applicability of the proposed AdaComp scheme, we tested it comprehensively on all 3 major kinds of neural networks: CNNs, DNNs, and LSTMs. For CNN, five popular networks for image classification were tested: MNIST-CNN, CIFAR10-CNN, AlexNet, ResNet18, and ResNet50. We also included in our tests two pure DNNs (BN50-DNN for speech and MNIST-DNN (not shown)), and an RNN (LSTM). In all these experiments we used the same hyper-parameters as the baseline (i.e., no compression). The selection of ${\rm L}_{\rm T}$ is empirical and is a balance between communication time and model accuracy; the same values are used across all models: ${\rm L}_{\rm T}$ is set to 50 for convolutional layers and to 500 for FC and LSTM layers. 

The experimental results are summarized in Table 2, while the detailed convergence curves are shown in Fig. 2. The proposed AdaComp scheme, on every single network (with different datasets, models and layers - CNNs, DNNs and LSTMs), tested under a wide range of distributed system settings (from 1 to 128 learners), achieved almost identical test errors compared with the non-compressed baseline. 

In addition to the conventional SGD with momentum optimizer, we also applied the AdaComp technique to other optimizers such as Adam \cite{kingma2014adam}. We ran experiments in Adam for the CIFAR10-CNN model and found that our compression scheme achieved similar compression ratios with Adam. In addition, the compression technique had no impact on model convergence or test error (Adam: baseline:18.1\% vs compression:18.3\%). As shown in Fig. 3, in comparison to SGD, as expected, Adam exhibited faster initial convergence but similar final accuracy. Intuitively, this result is consistent with the AdaComp algorithm which should be agnostic to the optimizer used (SGD with momentum vs. Adam vs. rmsprop) - with a detailed analysis presented in the next section.

Overall, our experimental results indicate that the AdaComp scheme is remarkably robust across application domains, layer types, learner numbers, and the choice of the optimizer. For the above ${\rm L}_{\rm T}$ choices of 50 and 500, the AdaComp algorithm typically selects only up to 5 elements within each bin (through sparsity). For ${\rm L}_{\rm T}$ sizes \textless  64, a sparse-indexed representation of 8-bits could be used effectively, while 16-bits of representation would be needed for larger ${\rm L}_{\rm T}$ sizes (up to 16K elements) - where 2-bits (out of 8 or 16) would be used to represent the ternarized data values. Therefore, in comparison with traditional 32-bit floating-point representations, the AdaComp scheme achieves an excellent \textbf{Effective Compression Rate} of $\mathbf{40\times}$ for convolutional layers and $\mathbf{200\times}$ for fully connected and recurrent layers.

\section{Discussions}
\subsection{Robustness of the AdaComp Technique} 
\begin{figure}[h]
  \centering
   \includegraphics[height=4.2cm]{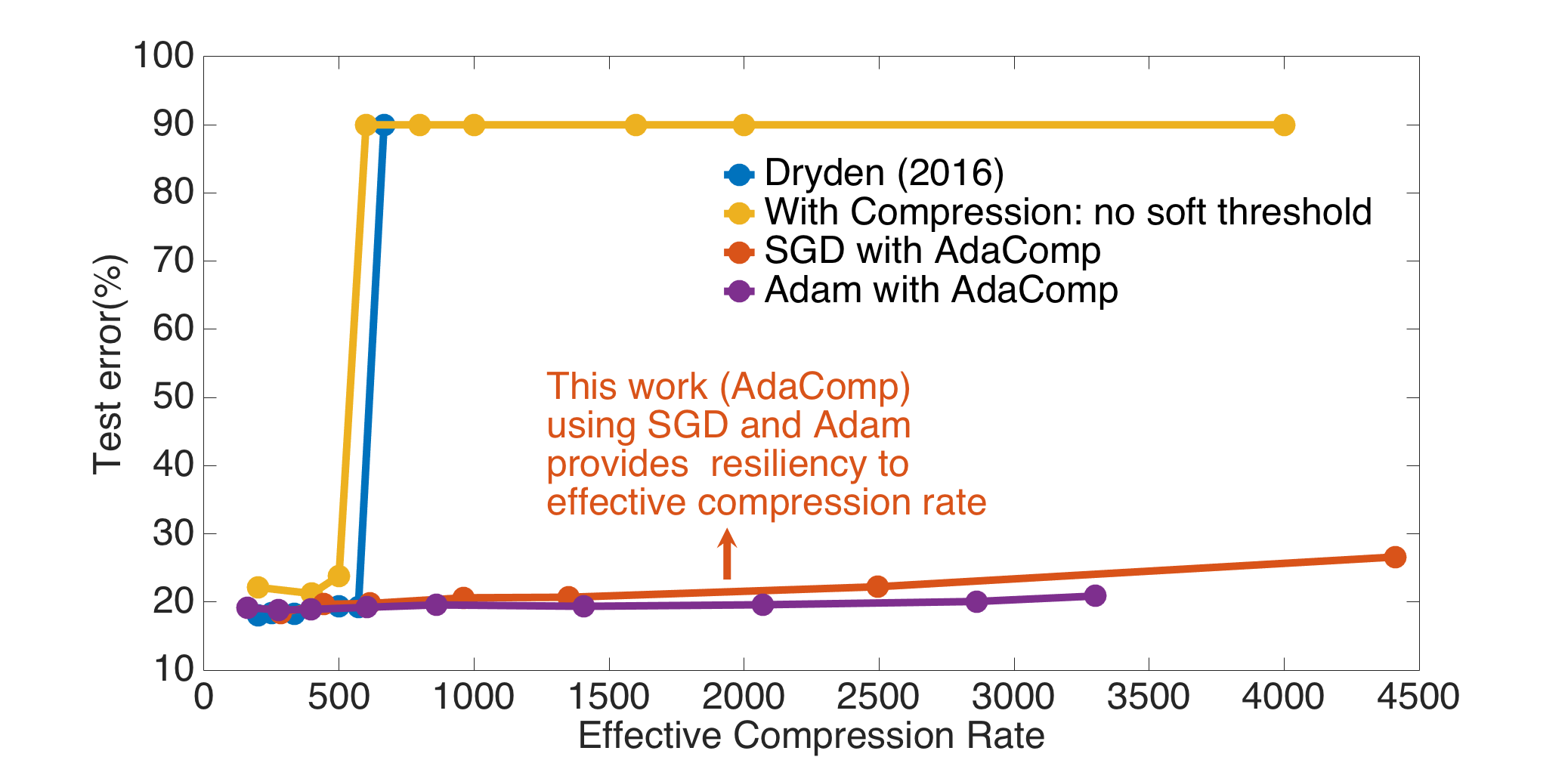}
  \caption{\label{fig:cifar10-cnn-comp-vs-testerr} Comparing CIFAR10 test-errors vs. effective compression rates for 3 different schemes - Dryden's method, Local Selection (LS) and the AdaComp technique trained by SGD. AdaComp is remarkably resilient to the compression rate while high compression rates cause LS and Dryden's schemes to diverge. AdaComp is also tested using Adam and exhibits even higher resiliency.}
\end{figure}

To understand the robustness of the different Residual Gradient compression schemes, 3 different methods are compared in Fig. \ref{fig:cifar10-cnn-comp-vs-testerr} - Dryden \cite{dryden2016communication}, Local Selection (LS), and this work (AdaComp). The LS technique refers to a scheme similar to AdaComp's local selection scheme, but without applying a soft-threshold to self-adjust the compression rate. This allows us to evaluate the importance of the self-adjustable nature of AdaComp. For each compression scheme, the CIFAR10-CNN model is trained using SGD with momentum (=0.9) and varying ${\rm L}_{\rm T}$ (and hence compression rates), reporting the final test error after 140 epochs. The 3 convolutional layers and the fully connected layer are all compressed at the same rate. As shown in Fig. \ref{fig:cifar10-cnn-comp-vs-testerr}, when the compression rate is less than 250, all methods achieve test errors close to (or slightly above) the baseline. While the test errors for Dryden's method and LS increase significantly as the effective compression rate increases, the AdaComp method is remarkably robust to ultra-high compression rates (reaching only 22\% test-errors for compression rates exceeding $2000\times$). We also show that this result extends to other optimizers including Adam - which shows even higher resilience (when compared to SGD) at high compression rates.

\begin{figure}[h]
  \centering
   \includegraphics[
    height=44mm
   ]{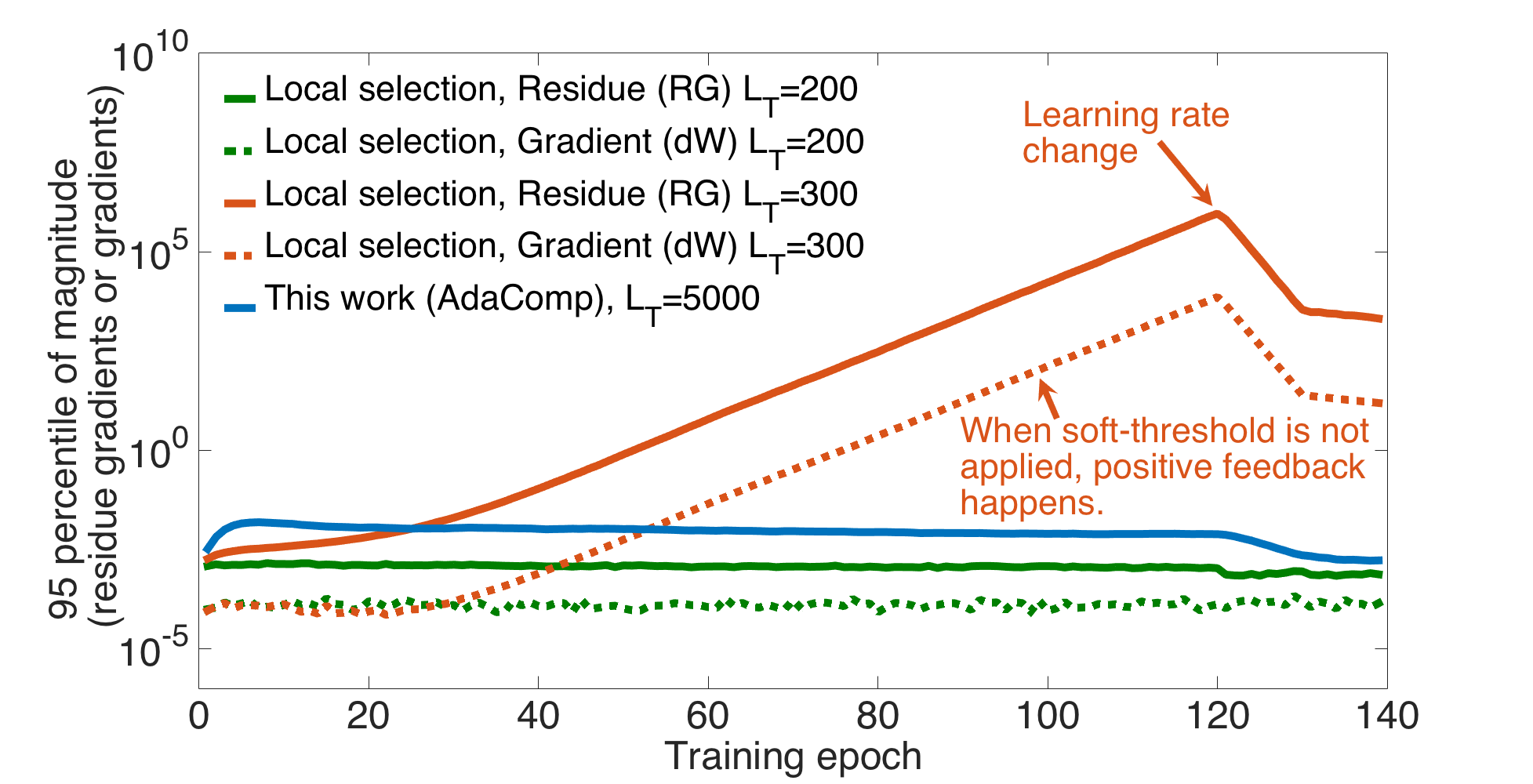}
  \caption{\label{fig:positive-feedback}Comparison of the magnitude of Residual Gradients (RG) for the local selection scheme (without adaptive soft-threshold) and the proposed AdaComp scheme. RG values are very sensitive to ${\rm L}_{\rm T}$ when local selection is used: larger ${\rm L}_{\rm T}$ (corresponding to higher compression rate) causes exponential increases in RG and results in model divergence. In comparison, AdaComp is very resilient to high ${\rm L}_{\rm T}$ values. }
\end{figure}

To understand why the LS scheme fails when the compression rate is high, we focus on experiments where the FC layer alone is compressed using LS. In Fig. \ref{fig:positive-feedback}, we plot the value of the 95 percentile of the gradient (dW) and the Residual Gradient (RG) during the training process. When the compression rate is small (e.g. LS with ${\rm L}_{\rm T}$=200), the magnitude of RG and dW is stable during training, leading to successful convergence with test errors (17.84\%) close to the baseline. However, as the compression rate increases (e.g., LS with ${\rm L}_{\rm T}$=300), the magnitude of RG and dW appears to increase exponentially, resulting in complete divergence. This exponential increase can be understood as a positive feedback effect - where an increase in RG results in higher training error, leading to further increase in dW. Since dW is accumulated into RG, this feedback loop further expedites the growth of RG. As a result, both RG and dW grow exponentially over epochs.

The key factor that makes the AdaComp technique robust is the self-adjustable threshold. The positive feedback in LS occurs because insufficient number of gradients are sent after each mini-batch. The difficult challenge with the LS and Dryden schemes is to find which and how many gradients need to be sent - since this number can be layer, network and hyper-parameter dependent. Recall that the AdaComp scheme sends a few additional residual gradients close to the local maximum in each bin - and can therefore automatically adapt to the number of important gradients in that set. Even if the compression rate is high, soft-threshold avoids the exponential increase in RG and dW - by being adaptive to the number of gradients being sent. This is demonstrated in Fig. \ref{fig:positive-feedback}, where RG for AdaComp with ${\rm L}_{\rm T}$=5000 slightly increases in the beginning, but stabilizes after that. Note that the compression rate using AdaComp (${\rm L}_{\rm T}$=5000) is ${\sim}2000\times$, which is much higher than ${\sim}600\times$ obtained with the LS scheme (${\rm L}_{\rm T}$=300).

\begin{figure}[h]
  \centering
   \includegraphics[width=0.7\linewidth]{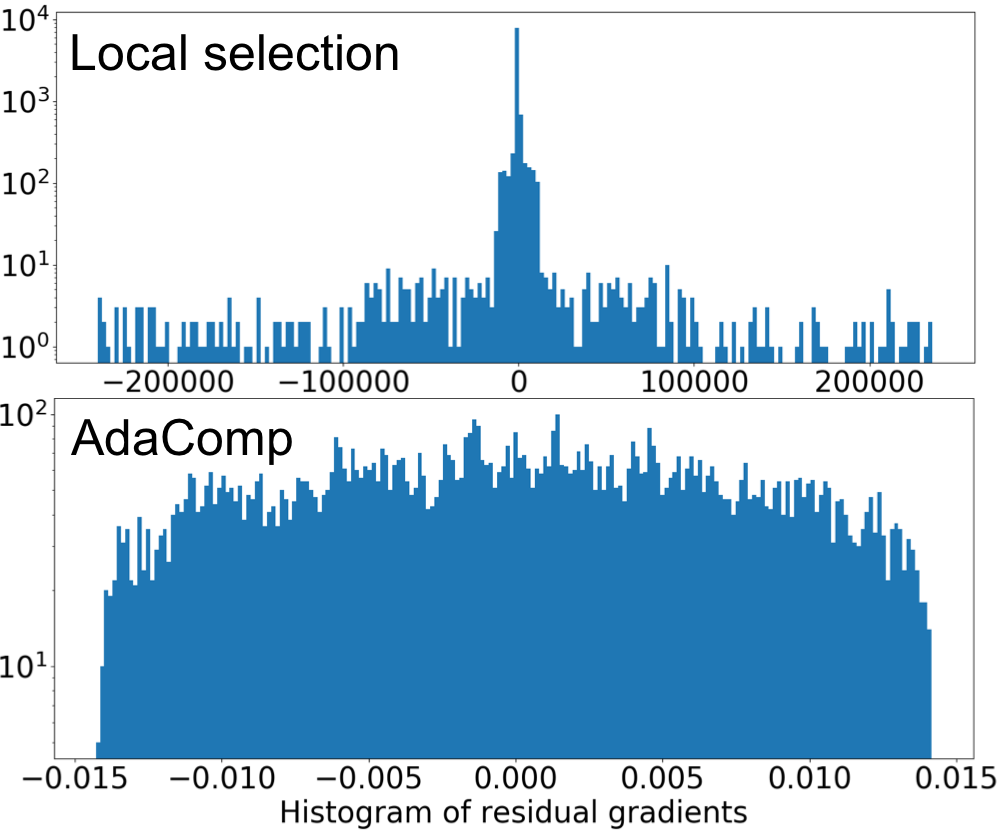}
  \caption{\label{fig:histogram_ls-vs-ll}Comparing histograms of the Residual Gradients (RG) at epoch 120: Local Selection (LS) technique (top) and the  AdaComp technique (bottom). AdaComp reduces RG by many orders of magnitude in comparison to LS.}
\end{figure}

To illustrate how the self-adjustable threshold impacts the magnitude of RG, we plot the histogram of RG for LS and AdaComp at epoch 120 in Fig. \ref{fig:histogram_ls-vs-ll}. For the LS scheme, only a finite number of gradients are sent, increasing the RG of the remaining gradients. Therefore, it is observed that RG exponentially increases (ranging from -240K to 239K), shaping the histogram to have a very long tail at both ends. However for the AdaComp scheme, the gradients with large RG are all sent altogether, drastically shortening the tails. Therefore, RG for AdaComp does not increase over epochs.

\subsection{Impact of mini-batch size and number of learners}

Empirically, as shown in Fig. 7(a), we observe that increasing the mini-batch size reduces the achievable compression rate (while preserving model fidelity); this is true for previous work \cite{dryden2016communication} as well as the AdaComp compression scheme. Recall that the advantage of AdaComp over Dryden's scheme is its ability to locally sample the residue vector and thereby effectively capture high activities in the input features. When the mini-batch size increases, we expect the input features at every layer to see higher activity. This causes the compression rate for Dryden's technique to suffer dramatically, because relatively small residues (or gradients) now become increasingly important. In contrast, AdaComp locally selects elements, so while the compression rate does degrade, it still does a far better job at capturing all of the important residues. This results in a ${\sim}5-10\times$ improved compression rate for AdaComp (Fig. 7(a)).

Consistent with the explanation above, we also observe (in Fig, 7(b)) that the compression rate for CIFAR10-CNN dramatically scales with the number of learners in the distributed system (while proportionally reducing the mini-batch size per learner). With more learners, each learner sees a smaller local mini-batch size and therefore compression rate is enhanced due to lower feature activity.

\begin{figure}[htb!]
        \centering
         \subfloat[Compression rate comparisons for different minibatch sizes.]{\includegraphics[ height=44mm]{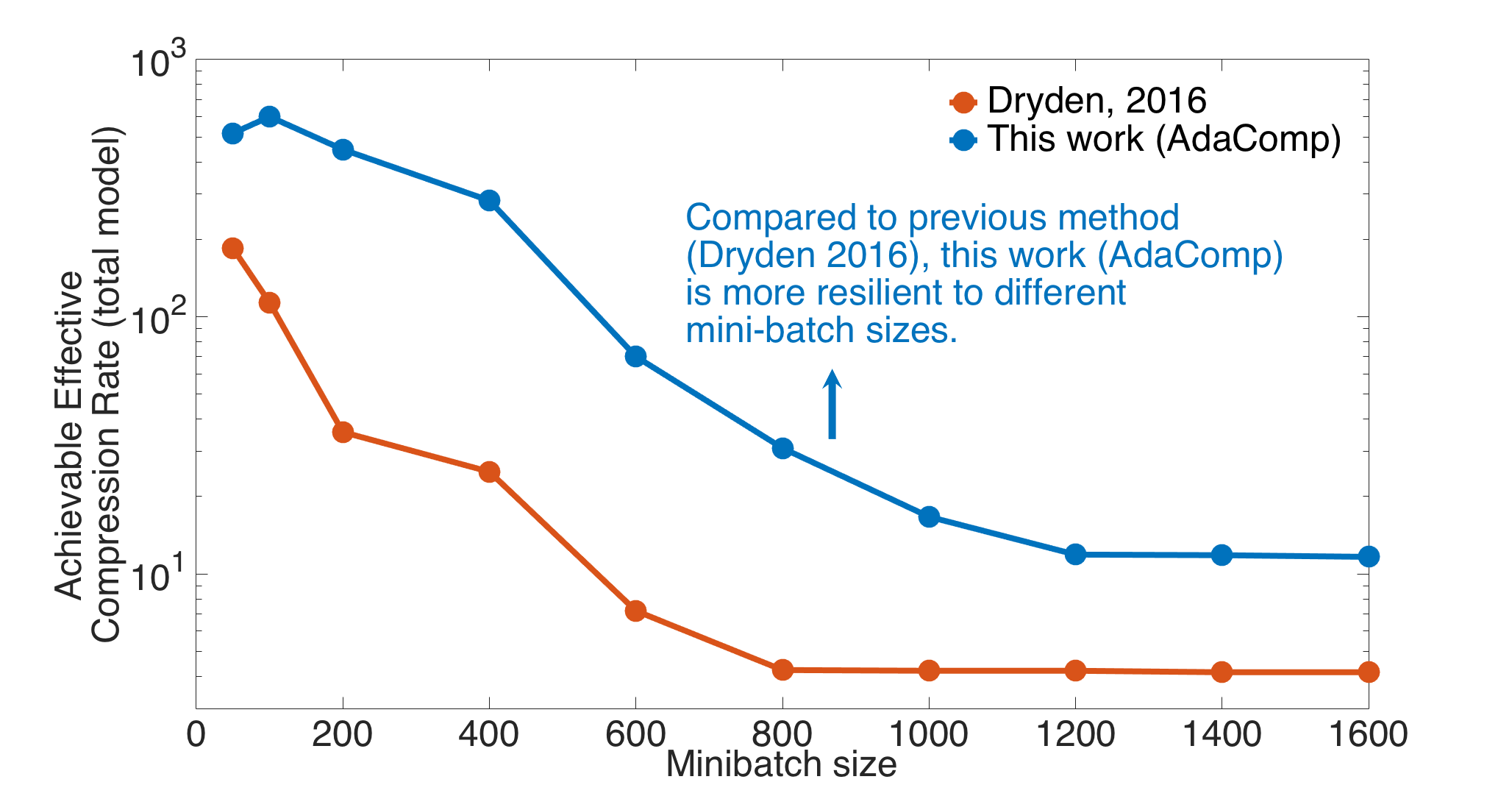}}
      
        \subfloat[Achievable compression rate for AdaComp with varying number of learners. Super-minibatch size (across learners) = 128.]{\includegraphics[ height=44mm]{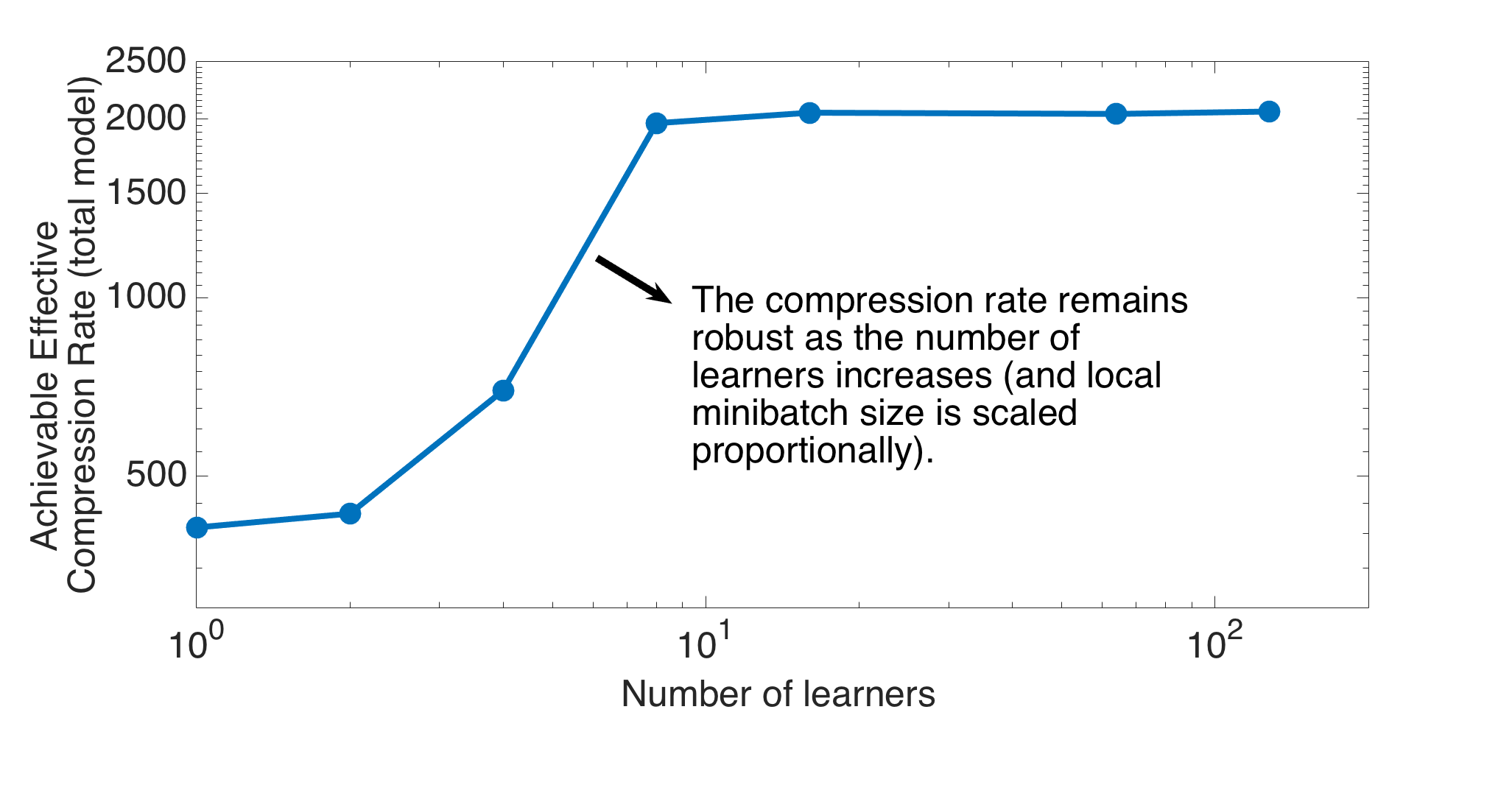}}
       
        \caption{\label{fig:mb-ln-comp-rate} Impact of mini-batch size and the number of learners on the compression rate for CIFAR10-CNN. Compared to previous work \cite{dryden2016communication}, AdaComp is more resilient to large minibatch size (a) and shows higher compression rate for large number of learners (b) - while maintaining test error degradation within the acceptable range (\textless 1\%).}
        \end{figure}

\section{Conclusions}
A new algorithm, Adaptive Residual Gradient Compression (\textbf{AdaComp}), for compressing gradients in distributed training of DL models, has been presented. The key insight behind AdaComp is that it is critical to consider both input feature activities as well as accumulated residual gradients for maximum compression. This is accomplished very effectively through a local sampling scheme combined with an adjustable (soft) threshold, which automatically handles variations in layers, mini-batches, epochs, optimizers, and distributed system parameters (i.e., number of learners). Unlike previously published compression schemes, our technique only needs 1 new hyper-parameter, is more robust, and allows us to simultaneously compress different types of layers in a deep neural network. Exploiting both sparsity and quantization, our results demonstrate a significant improvement in end-to-end compression rates: ${\sim}200\times$ for fully-connected and recurrent layers, and ${\sim}40\times$ for convolutional layers; all without any noticeable degradation in test accuracy.
These compression techniques will be foundational as we move towards an era where the communication bottlenecks in distributed systems get exacerbated due to the availability of specialized high-performance hardware for DL training.

\section{Acknowledgments}
The authors would like to thank Naigang Wang, Vijayalakshmi Srinivasan, Swagath Venkataramani, Pritish Narayanan, and I-Hsin Chung for helpful discussions and supports. This research was supported by IBM Research AI, IBM SoftLayer, and IBM Congnitive Computing Cluster (CCC).

\bibliography{aaai2018}
\bibliographystyle{aaai}

\end{document}